# A multiagent urban traffic simulation
# Part I: dealing with the ordinary

Pierrick Tranouez, Patrice Langlois, Éric Daudé

*Abstract*— We describe in this article a multiagent urban traffic simulation, as we believe individual-based modeling is necessary to encompass the complex influence the actions of an individual vehicle can have on the overall flow of vehicles. We first describe how we build a graph description of the network from purely geometric data, ESRI shapefiles. We then explain how we include traffic related data to this graph. We go on after that with the model of the vehicle agents: origin and destination, driving behavior, multiple lanes, crossroads, and interactions with the other vehicles in day-to-day, "ordinary" traffic. We conclude with the presentation of the resulting simulation of this model on the Rouen agglomeration.

## I. INTRODUCTION

IF traffic modeling is nearing a century of age, most of these models belong to Operational Research problems – finding an optimal solution balancing various constraints. In these models, roads and road users were abstracted and aggregated, so as to become a flow problem that could then be optimized. They can answer interesting questions in urban or public transport planning [1].

Sometimes considering average response to a problem is not enough for the scientific problem at hand. We are interested in a dynamic modeling of urban traffic. In this kind of problem, the actions of a few can have a definite impact on the global traffic. An accident implicating half a dozen vehicles in a strategic crossroads of a town can create a traffic jam wave that can affect thousand vehicles. This is the kind of complex phenomenon we would like to be able to model and simulate. Classic OR tools aren't well suited to the task.

Although we aren't the first to make this statement [2,3], models that tried to alleviate this too-large-scale limitation, have mainly tried to use cellular automata for the task. They added some level of individual-based components to their modeling, but still failed to encompass all that could be needed. Cellular automata are eulerian methods – intelligence is in one place, rules describe the behavior of bits of space.

The authors would like to thank the GRR SER and the region Haute-Normandie for the funding of the MOSAIIC program from which this work stems from.
P. Tranouez is with Litis, Rouen University. (e-mail: Pierrick.Tranouez@univ-rouen.fr)
Patrice Langlois and Éric Daudé are with UMR IDEES, Rouen University (email: Patrice.Langlois@univ-rouen.fr and Eric.Daude@univ-rouen.fr)

Values linked to the cells seem to simulate the entities of the modeled system, the same way alternatively lit crystals in an LCD display can give the illusion an object moves around a screen. This contrasts with lagrangian descriptions, where entities of an environment are distinguished, and their spatial coordinates are but one of their describing characteristics. Unlike what can be easily simulated in a CA, lagrangian entities have a *trajectory*: even in a discretized space a la CA, they can for example act according to something that happened $n$ time steps and $m$ space steps before or away, or according to a plan. This can't *practically*[1] be done in a CA. Multiagent systems belong to this latter category of modeling. As we try to build a model with a grain fine to the level of geometrically correct individual vehicle behavior, from which at least town-quarters-level flow disturbance can arise, we believe this technique is the right one for the task.

## II. FROM GEOMETRY TO TOPOLOGY

### A. Geographical databases

A Geographical Information System is a system designed for creating, storing, analyzing and managing spatial data and associated attributes. Although it contains a relational database, it needs to go beyond what is needed for classical alphanumerical databases to manage geometrical information, which is continuous by nature, as opposed to the discreteness of usual databases. Indeed for example the database cannot contain all the points of two segments in order to compute a possible intersection: other storing and managing methods must be used for the geometric data of the system.

A geographic database is generally comprised of layers or coverage overlapping on a same spatial domain. Each layer contains homogeneous spatial features such as the limits of a city, the course of a river, the geometry of a road etc. Each feature is described in two different ways. First the geometric and optionally topological information is stored in different binary files in the base. Second the record description is a line in the record table; it contains different attributes and descriptions of the feature (generally text or numbers).

### B. ESRI shapefiles

The first step of the constitution of our system is the constitution of a basic layer of geographic database. This layer is built from the importation of *shapefiles*, a GIS file format popularized by ESRI [4]. In order to build a traffic simulation, we will build our model from data relative to the road network and optionally from other localized information such as living or working areas.

---
[1] As opposed to theoretically, as cellular automata are universal calculators

A shapefile is mainly constituted of three files: one contains the attribute table (.dbf), another contains the geometric data (.shp) and the third is an index allowing matching entries of the first with those of the second.

A shapefile contains only the geometric description of objects through a collection of 2D or 3D coordinates that represents, according to the layer type, a cloud of points, open polygon lines (for networks or closed polygon lines to describe the boundary of surfaces. The topological information, which describes in geomatics the relationships between the geometric entities, such as connections of edges with nodes in a graph or the adjacency between zones in a surface partition, is absolutely not present in a shapefile, and must therefore be computed by our application form the raw geometry of the imported data.

To build a realistic representation of the traffic network of an important urban agglomeration able to simulate the circulation of tens of thousand of vehicles, we had to conceive a network layer structure both complex and efficient. Furthermore, the importation of data coming from existing data provider such as IGN, *NAVTEQ* or Tele Atlas, we had to deal with the way each modeled things in their solutions.

### C. Urban network structure specifications

A road network is modeled according to specifications that are in part common to any network and in part dependent on decisions made by the data producer.

#### 1) General specifications

A road network shares the properties of any geographic network. It is constituted of two main geometric entities: *lines*, linear components, comprised of several shape points, and *nodes*, point components that join or terminate lines.

These two entities are joined in an *oriented multigraph* $G = (S, A, f)$ where S is the set of vertices, associated to the geometric nodes, while the set A of edges is associated to the geometric lines, while function $f : A \to S \times S$ associates to each edge one initial and one final vertex.

Unlike most other geographic information layer, a road network may not be planar: two lines can intersect in their planar projection without modeling an intersection in the real world. This happens when these lines are at different altitudes such as in bridges, tunnels, or motorway embranchments.

Furthermore, geographic graph are *topological graph* differ from usual graph in that they are associated to one geometric representation, called the *embedding* of the graph. Only vertices of degree 3 or more are considered to be true vertex, those of degree 2 being seen as *shape points*, useful for the geometric information they bring, but not "true" connectors. The geometric representation of the graph is always present to the mind of the geographer, which may create misunderstandings with other scientists more used to a more abstract representation of graph, with *planar* graph rather than *plane* graph. As previously said, it is also sometimes extended to non planar graph: the geometric information in the shapefile represents in that case the projection on a connected compact 2-manifold of a graph embedded in a connected compact 3-manifold (intuitively: a 3D graph is drawn on a surface).

The attribute table associated to the network will contain all the traffic related information, such as the number of lanes, speed limits, sense of travel etc. Nonetheless this information may not be associated to elementary lines or nodes. For example major roads may contain different lines and important roundabout may contain different nodes and lines. We therefore defined the notion of super-nodes that relate to several nodes (and the assorted sub-graph) and super-edges that relate to several edges (and the assorted sub-graph). G is therefore a hypergraph in these conditions. Whether these are met or not depend on modeling decisions made by the data provider.

#### 2) Geographic data based specifications

There are different ways of structuring the geographic information in a shapefile to model a network.

For example *NAVTEQ* chose in its Navstreets product to create a node for each intersecting link, even if the road they model are not connected. Another layer represents the relative elevation of the entities of this first layer. Both must therefore be used to correctly build the road network in our simulation. Another example is the orientation of the edges, as the links are oriented following another convention (called "Reference nodes") than what could be used in a shapefile, and the edge must therefore be computed following this convention.

### D. Building the topology from the geometry.

Building a topology from the geometric information contained for example in a shapefile depends on the kind of spatial organization we want to represent.

#### 1) Planar mesh

In the case of a surface mesh (ex: limits of countries, of urban areas, of town quarters etc.), we aim at rebuilding the boundaries and the junction nodes between them from closed polygonal chains (aka polylines). The layer we produce is thereafter structured around a planar multigraph of vertices, edges and faces, and with each oriented edge associated to 2 vertices (initial and terminal) and to 2 faces (left and right).

The building algorithm uses a quadtree and a tree connecting each point, in which all the points of the shapefile are organized. Each leaf of the quadtree contains a point $P_i$ and 4 branches for the 4 quadrants of space (NE, SE, SW, NW) surrounding $P_i$. When a branch is a leaf, it contains a point belonging to right quadrant relatively to its father, and vice versa. This structure allows for a quick detection of the multiplicity of points. For example, a point with a multiplicity of 3 or more will be associated to a vertex, while a point of multiplicity of 2 will be a shape point of an edge. Furthermore, the connection tree allows the quick detection of adjacent points along a polyline, and detecting the superposition of two lines forming the boundaries of two zones, or the succession of angular sectors around a vertex common to three polygons or more.

*2) Network*

In order to build the structure of a planar network (for example hydrographic or of roads), we do not store faces but the polar order of succession in the edges. Each edge stores the next edge turning left and the prior edge turning right. This structure is known as DCEL, Doubly Connected Edge List [5]. The algorithm to generate this topology uses the same dynamic quadtree structure to build the DCEL.

The road network often exists in 3D, although despite the existence of this possibility, most shapefiles only contain a 2D geometric representation. The data provider must in that case model the altitude differently, and our algorithm must be adapted to this. For example *NAVTEQ*'s Navstreets [6] uses another layer called z-levels that must be consulted to know whether a point corresponds to a node or not.

At the end of this step, we have a topological graph that is structured like the road network, but without its semantics. We will now build from it and from the database part of the shapefile a non-topological graph that models this ontology.

*E. From static topology to traffic-oriented network*

*1) Traffic oriented graph*

Our traffic model is individual-based: each vehicle will be modeled as an agent. This implies the creation of an adapted environment for them, in terms both suitable to their ontology and adapted to the geographic data we reaped. For that a graph will be built, a *transport graph* that will contain the necessary structures and values.

This first version of our models is only interested in simulating motor vehicle: pedestrians and bicycle are ignored.

The database contains the sense of travel and the traffic restrictions for each topological edge. One oriented edge is created for each sense of direction allowed for motor vehicles.

Edges and vertices of the transport graph are called *elements*. To each element is associated a *data container* and a *vehicle transporter*.

The *data* associated to an edge are for example its geometric length, its number of lanes, its speed limits etc.

The data associated to a vertex is notably the size of the container of its transporter, depending on the number and the sizes of the edges connected to him.

*Transporters* are non-mobile agents associated to elements. They handle parts of the collective behavior of the vehicles. They will be described in more depth in the following section.

*2) Routes in the graph*

Mobile agents will try to reach destinations in the graph. As we intend to simulate a realistic traffic of tens of thousand of vehicles, we want to facilitate their computing of their trajectory. To do that, we build a set of "shortest" path stored in the traffic graph.

We compute a weight on the edges that combines different parts of its data: its length, the speed it can reasonably be driven upon, its estimated width based on the number of lanes etc. to model the attractiveness of this edge. After that we compute Dijkstra's algorithms [7] from each vertex to all the others, which we store in each vertex. This data takes (numberOfVertices)$^2$ bytes of data, which is important, but allows the computation of a good path by an agent in constant time, which is a good thing as hundreds of agents are generated at all time in the simulation (simulating vehicles entering the road network of the simulated urban agglomeration).

III. MOBILE AGENTS OF THE NETWORK

Our agents are mainly so far car agents, trying to go from one place to another.

*A. Strategic behavior*

Modeling in details the various detailed trajectories of car users is a research problem in itself [8]. Nonetheless we are not interested in who did what or why, but only in what are the fluxes in our network in typical scenarios. When an agent is injected in the network, a starting point and a destination are randomly chosen.

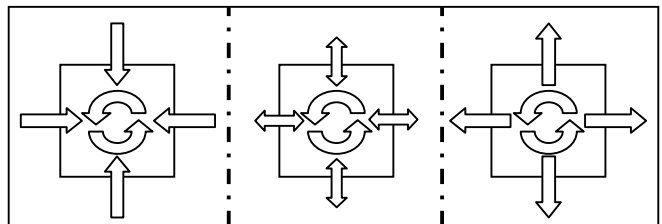

Fig. 1. 3 different scenarios of source/destination pseudo-random choice

This randomness is not necessarily uniform. If we suppose the agglomeration centered on its main town, like the agglomeration of Rouen that we simulated more than others, we can shape different distribution, favoring the likelihood of drawing rather a inner or an outer edge for example. Traffic between 8:00 AM and 9:00 AM for example starts mainly on the border or outside the agglomeration and ends to the same distance to the center (outer edges): we can simulate traffic that do that. When shops close in the town center, we have a traffic that is mainly outer bound, with a more important center generation: we can simulate that. We do not have to know what this car and its driver did in the morning, we don't have to simulate realistically its history, as long as we model the actual traffic fluxes right.

Once the agent knows where it is, and what its destination is, it can use the best paths stored in the traffic network to plan a trajectory. It then drives here, adapting his path through its tactical behavior, and managing its immediate surroundings through its operational behavior.

*B. Operational behavior*

The planned trajectory of an agent is a succession of edges. Once in an edge the agent tries to drive to its end, the next connection, where it will be able to choose the next planned edge.

When it enters an edge, the agent first chooses a lane if several are available, based on the traffic density in each, with a bias for the rightmost lane. As we have a good geometric description of the lane, the driving behavior is fairly detailed, incorporating the length of the car, its capacity/will to accelerate and brake, the taste of its driver for long/short safety distance, its taste for following or breaking speed limits etc. All this is incorporated in a driving model inspired by Martin Treiber's Intelligent Driver Model [9]. IDM is a longitudinal traffic model, so we had to expand it to handle multiple lanes and crossroads – the original IDM works for an unlimited one-way, one-lane road – we did not use Treiber's MOBIL lane changing model as it is better adapted to motorways than to urban lane changing decisions.

The data provided by geographic providers does not include right of passage or traffic lights at crossroads. We therefore had to develop our own model aiming at the simulation of crossroads in a heavy traffic.

When a vehicle reaches a crossroad, it slows down and acts according to the fluidity of traffic in the crossroad, in the edge it is currently upon and in the edge it whishes to go to. If they are encumbered, it will more often wait in its way, but it may enter the crossroad and wait here, thus encumbering it (with a more or less strong individual tendency to do so). If the edge it is aiming at has multiple lanes, it will watch both of them, to see if it could fit in one.

### C. Tactical behavior

Although vehicles have an original plan, they will adapt it to what they perceive of their environment. When stuck in what they perceive is a jam, they will try to find alternate routes out of it to their destination.

The first method we used is the simpler one. When a vehicle doesn't move enough to its liking – this saturation is variable amongst agents – it tries to take alternate paths as soon as possible, favoring the roads with least dense circulation – although this is not absolute, so as to avoid loops. Once it estimates it's far enough from the jam that sprang this alternate behavior, it resumes using the best path table to find a suitable one to its destination.

The second one is more sophisticated, as it will have uses beyond mere traffic avoidance. Its intelligence is modeled more in the Transporter agents than in the vehicles. Transporters estimate their encumbrance. To do that, they employ direct measure – how many vehicles do they contain over how many vehicles can they contain in average – but also statistics on the proportion of vehicles they contain that are annoyed by the traffic – as described in the first method – and information from the Transporters around them. If based on this they decide they are *encumbered* they also warn the Transporters around them of their perception. This will lower the threshold for them to feel encumbered.

Once *encumbered*, the nodes they are connected to will recompute their best path table, using a huge weight for the encumbered edges. When a vehicle arrives to one of these nodes and wants to go to one of the jammed edges, it is informed of the edge state, and it can recompute a route around it, or take the edge anyway.

This mechanism is also theoretically interesting, as it is an implementation of an emergent property: the interactions of individual behavior affect the behavior of an agent of an higher scale, who alters his behavior, which in turns transforms the behavior of the lower level vehicles. This reifies the perception an individual driver can have of the state and dynamics of the traffic he is plunged in as a whole.

A Transporter can also be *barred*, because of an accident for example. In that case the same mechanism is used, except that this time circumnavigating is mandatory.

The mechanism of these two states is especially useful in what was the original purpose of our model and its main application: simulating urban important accidents – such as industrial accident – as the modeler can bar the edges it wants as part of his scenario, and see how the traffic adapts to it in simulation real time, as the vehicles discover the evolving road network and fluxes. This is the application that will be developed in the part 2 of our article.

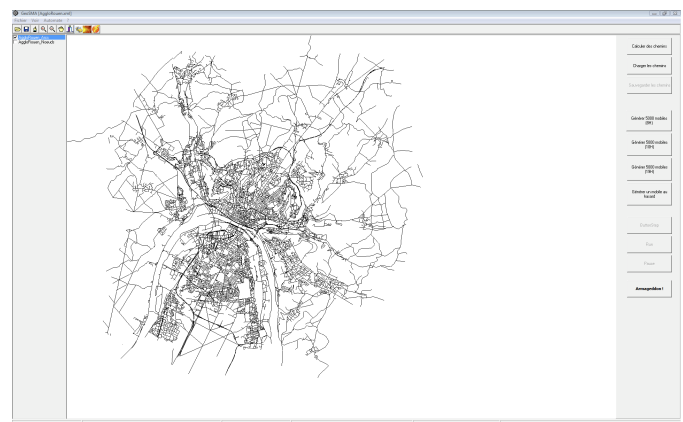
Fig. 2. A screenshot of our simulator, loaded with Rouen agglomeration

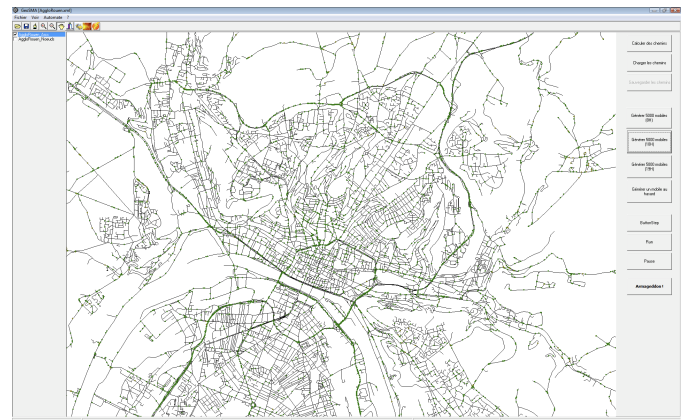
Fig. 3. In green vehicles in an example of running traffic, in a zoom on Rouen itself.

### IV. CONCLUSION

This article is more technical than thematic. We tried to write the kind of article we would have liked to read when we started on this work. We have nonetheless done thematic validation.

One of the problems for the validation is that modeling as seldom been taken to such a detail level. This level is necessary because of the multi-level nature of traffic: the decision of one driver can start a jam or jams for thousands of drivers, half a town away, half an hour later, a la butterfly effect. Macro model of fluxes, which dominate the field of traffic simulation, cannot do that. Their validation for example is often based on the fundamental diagram of traffic flow of a few selected axes, for hourly traffic. We can compute second by second fundamental diagrams of each edge of our network. We can therefore be an order or two of magnitude more precise in our measure, but what to do of all this information? Indeed, if we describe in details the behavior of vehicles, one must not lose sight that they are *not* what we are trying to model, the traffic is what we are trying to model, that is their behavior as a group. We fine-tune individual behavior to have the emerging group behavior right.

What we have ascertained so far is that:
- Most edges comply with the Fundamental Diagram made over 5 minutes of time, most of the time. This remains the case even once the measure-time unused edges are taken out of the count
- We simulated our university home agglomeration of Rouen with up to 50 000 vehicles, and traffic specialists find the results subjectively very satisfying

We tried to compare the results of our simulations with data we had about the traffic of the Rouen agglomeration. The data dated from 2001, while the geographic data we had for the network dated from 2006-2007. The western part of the road network had changed too much during this period for any solid conclusions to be drawn from it, despite superficial resemblances in other parts of the networks. We have contacted the road management of the agglomeration for more recent data.

Finally, as we will describe in further details in part 2 of this article, we have a toolbox to simulate urban industrial accident and its effect on traffic with a realistic level of traffic volume.

REFERENCES

[1] May, Adolf. Traffic Flow Fundamentals. Prentice Hall, Englewood Cliffs, NJ, 1990.
[2] http://transims-opensource.org/
[3] http://www.matsim.org/
[4] ESRI Shapefile Technical Description, ESRI White Paper, July 1998
[5] F. P. Preparata, M.I. Shamos - *Computational Geometry: An introduction*- Springer Verlag, 1985
[6] *NAVTEQ*'s NAVSTREETS Street Data Reference Manual v2.0 -- 16 June 2006
[7] E. W. Dijkstra: "A note on two problems in connexion with graphs". In Numerische Mathematik, 1 (1959), S. 269–271
[8] A. Banos, T. Thévenin  (2008) "Création de champs de potentiel et simulation d'itinéraires à partir de l'enquête ménages-déplacement"  In: *Information Géographique et Dynamique Urbaine*, Volume 1 : Analyse et simulation de la mobilité des personnes  Edited by:Marius Thériault et François Des Rosiers.   Hermes
[9] A. Kesting, M. Treiber, and D. Helbing (2007) "General lane-changing model MOBIL for car-following models" *Transportation Research Record: Journal of the Transportation Research Board*, Volume 1999/2007, pp. 86-94. DOI 10.3141/1999-10